\documentclass[conference, 12pt]{IEEEtran}
\usepackage{amsfonts}
\usepackage{amsmath}

\usepackage{graphicx,times,amsmath}

\IEEEoverridecommandlockouts    

\begin{document}
\title{ \LARGE\bf Deep Learning for Skin Lesion Classification }

\author{
    \IEEEauthorblockN{P. Mirunalini\IEEEauthorrefmark{2}, Aravindan Chandrabose \IEEEauthorrefmark{2}, Vignesh Gokul \IEEEauthorrefmark{2}, S. M. Jaisakthi\IEEEauthorrefmark{1}}
 
    \IEEEauthorblockA{\IEEEauthorrefmark{2}Department of Computer Science \& Engineering, SSN College of Engineering, Chennai, India}    
    \{miruna,aravindanc\}@ssn.edu.in, {vigneshgokulgv}@gmail.com,

   \IEEEauthorblockA{\IEEEauthorrefmark{1}School of Computing Sceince \& Engineering, VIT University, Vellore, India}  
   {jaisakthi.murugaiyan}@vit.ac.in}

\maketitle

\begin{abstract}

Melanoma, a malignant form of skin cancer is very threatening to life. Diagnosis of melanoma at an earlier stage is highly needed as it has a very high cure rate. Benign and malignant forms of skin cancer can be detected by analyzing the lesions present on the surface of the skin using dermoscopic images.  In this work, an automated skin lesion detection system has been developed which learns the representation of the image using Google's pretrained CNN model known as Inception-v3 \cite{cnn}. After obtaining the representation vector for our input dermoscopic images we have trained two layer  feed forward neural network to classify the images  as malignant or benign. The system also classifies the images based on the cause of the cancer either due to melanocytic or non-melanocytic cells using a different neural network. These classification tasks are part of the challenge organized by International Skin Imaging Collaboration (ISIC) 2017. Our system learns to  classify the images based on the model built using the training images given in the challenge and the experimental results were evaluated using validation and test sets. Our system has achieved an overall accuracy of 65.8\% for the validation set.
\end{abstract}

\section{Introduction}
Skin protects the body against infection and injury and helps to regulate body temperature. Cancer cells are formed on the skin when healthy cells change and grow uncontrollably, forming a mass called tumor. A tumor can be malignant or benign. Malignant tumor is very dangerous and can grow and spread to other parts of the body. A benign tumor may or may not be cancerous and it may grow but will not spread to other parts of the body. The benign or malignant tumors may arise from either melanocytes or non-melanocytic cells of skin.

Melanoma is a malignant tumor of melanocytes and it develops when unrepairable DNA damage has happened to skin cells which triggers mutations. These mutations cause the skin cells to multiply rapidly and form malignant tumors. These tumors originate in the pigment producing melanocytes in the basal layer of the epidermis. The majority of melanomas are black or brown, but they can also be skin-colored, pink, red, purple, blue or white. Melanoma is caused mainly by intense, occasional UV exposure (frequently leading to sunburn), especially in those who are genetically predisposed to the disease.  Melanocytic Nevus \cite{Nevus} are benign proliferation of melanocytes. They usually present as a brownish lesion with well defined borders with excess hair in it. The surface of the nevus may  be roughed. Seborrheic keratoses \cite{keratosis} are common, benign, pigmented epidermal tumors.  Lesions appear as coin-like, sharply demarcated, exophytic lesions and are “stuck on the skin” with a verrucous, rough, dull or punched-out surface. Flat lesions often have a smooth surface and are scarcely elevated above the surface of the skin. If melanoma is detected early and treated, it can be cured, but if not, the cancer spreads to other part of the body and can be fatal.  

A statistical analysis  of color features on two different feature space such as lesion and object feature space has been proposed by Cheng Y et al.  \cite{color} to classify various type of skin lesions. Chang W-Y and Huang A and Yang C-Y et al \cite{CAD} proposed a Computer Aided diagnosis system which used conventional and new color-related image features to classify the lesions as benign or malignant using support vector machines (SVMs). The review of the state of art of different systems and current practices, problems, and prospects of image acquisition, pre-processing, segmentation, feature extraction and selection, and classification of dermoscopic images were reported in \cite{review}. The classification system of skin lesion involves preprocessing the image for noise and hair removal, segmentation of lesions and extracting features from the lesions to further classify them.  Instead of going for explicit feature engineering, techniques have been developed which learn an image representation using deep neural networks.  

Our proposed automated system aims to detect the type and cause of the cancer directly using image representation obtained from Google's Inception-v3 model. Based on the representation vector we have performed two phase classification using 2 layer feed forward neural network with softmax activation function in our output layer. The two phase classification is achieved using two different neural networks with same representation vector. In phase-1 we have determined the type of cancer  either as malignant and  benign and the phase-2 involves determining the cause of cancer either due to melanocytic or non-melanocytic cells.

\section{ Proposed Methodology}
\subsection{System Architecture}
\begin{figure}[!h]
\centering
\includegraphics[width=3.5in]{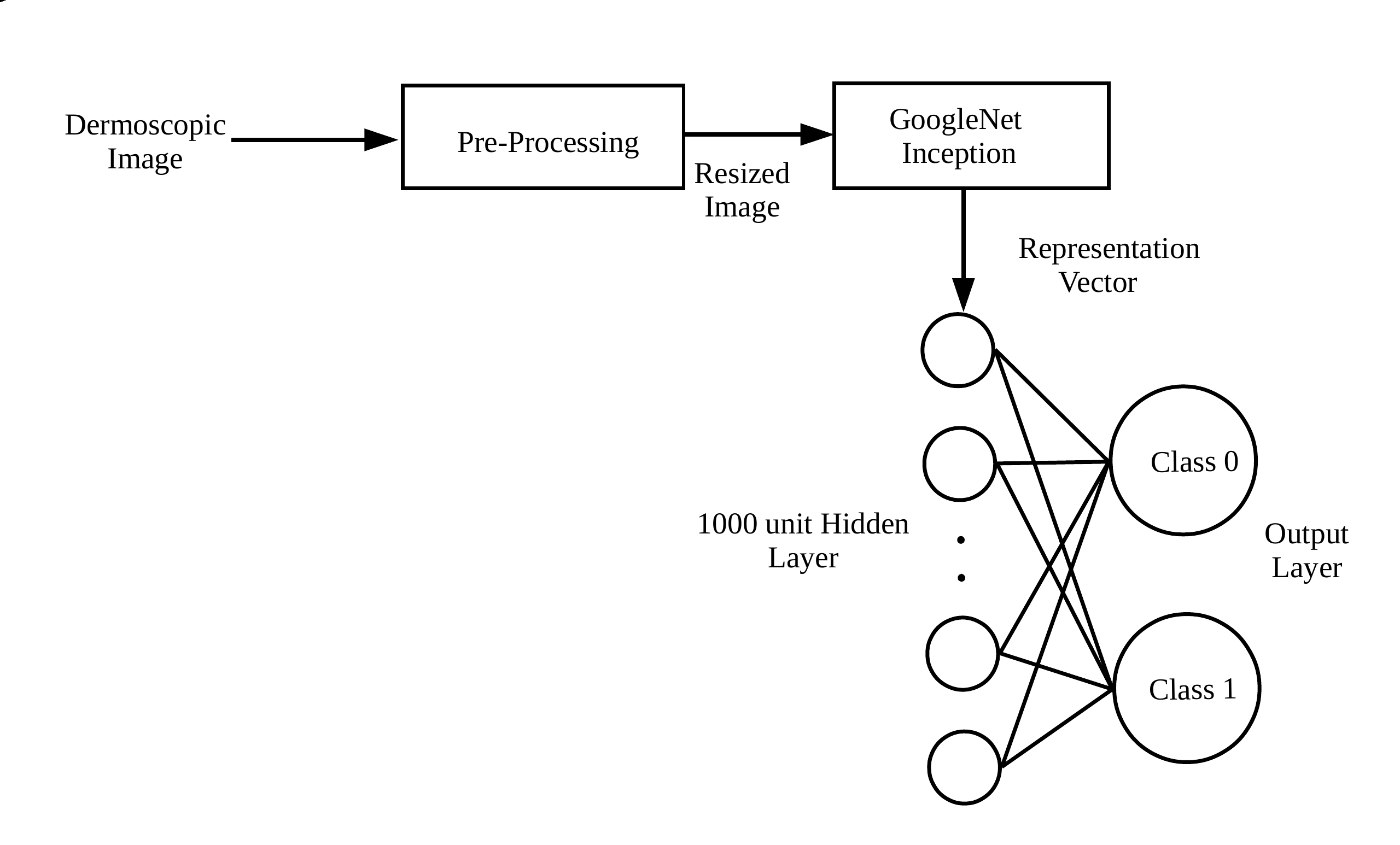}
\caption{System Architecture}
\label{fig1}
\end{figure}

The system takes input as dermoscopic images of various sizes. The overall training process consists of three stages: Preprocessing, Representation Learning and Classification. The system architecture is depicted in the Figure \ref{fig1} 

\subsubsection{Pre-Processing}
The images are first preprocessed and resized to the size of 299x299x3 so that it can be fed to the inception module. Many preprocessing techniques were applied to improve the test accuracy. From the traning set 20\% of images were chosen randomly  and those images  were mirrored, cropped, scaled and increased in brightness 
These preprocessed images were included  in the training set.

\subsubsection{Representation Learning} 
The inception module consists of multiple convolutional and pooling layers. The convolution layers consists of filters of size 7x7, 3x3, 5x5, 1x1 and of varying strides 1 and 2. Similarly, the architecture consists of both max pooling and average pooling of size 3x3 and 7x7 respectively with stride 2. To effectively compute deep convolutions, it makes use of various inception modules in between which consists of smaller 1x1 and 3x3 convolutional and pooling layers. Each convolution filter convolves over its input computes the weighted sum. 

Pooling layers are used to reduce the size of the image and also provides translation invariance. Pooling is achieved by using filters and pooling strategies such as max pooling, mean pooling and min pooling.

The representation  vector of the input image is obtained at different hierarchy level using the different layers of convolutional neural networks. The 1000 dimensional representation vector obtained at the next-to-last layer of the Inception-v3 model is used as input to our 2-layer feed-forward neural network.

\subsubsection{Classification} 
The two layer feed forward neural network consists of 1000 units in the hidden layer and 2 units in the output layer. The input to our network is 1000 dimensional representation vector of the image obtained from the fine tuned Google's  inception model. The output of hidden layer is fed to the softmax layer which outputs the probabilities of the image belonging to each of the class. Since for both the classification task, the images are the same, we use the same representation learned by the inception module and feed it to two different feed-forward neural networks. We use the Adam optimizer and cross-entropy loss for updating the weights by backpropagating the error backwards. The system was trained for 4000 iterations to obtain weights for the classification.

\section{Results and Discussion}
\subsection{Data Set}
The training dataset consists of 2000 dermoscopic images of skin lesions in JPEG format along with ground truth labels. The validation set has 150 images and the testing set consists of 600 images. The ground truth contains the image id, for phase-1 classification binary value of 0 corresponds to  benign and 1 for malignant  type of cancerand in phase-2 classification value of 1 corresponds to non-melanocytic cells and 0 for melanocytic cells.

\subsection{Result}
\begin{figure}[!h]
 \begin{center}
\includegraphics[width=3.2in]{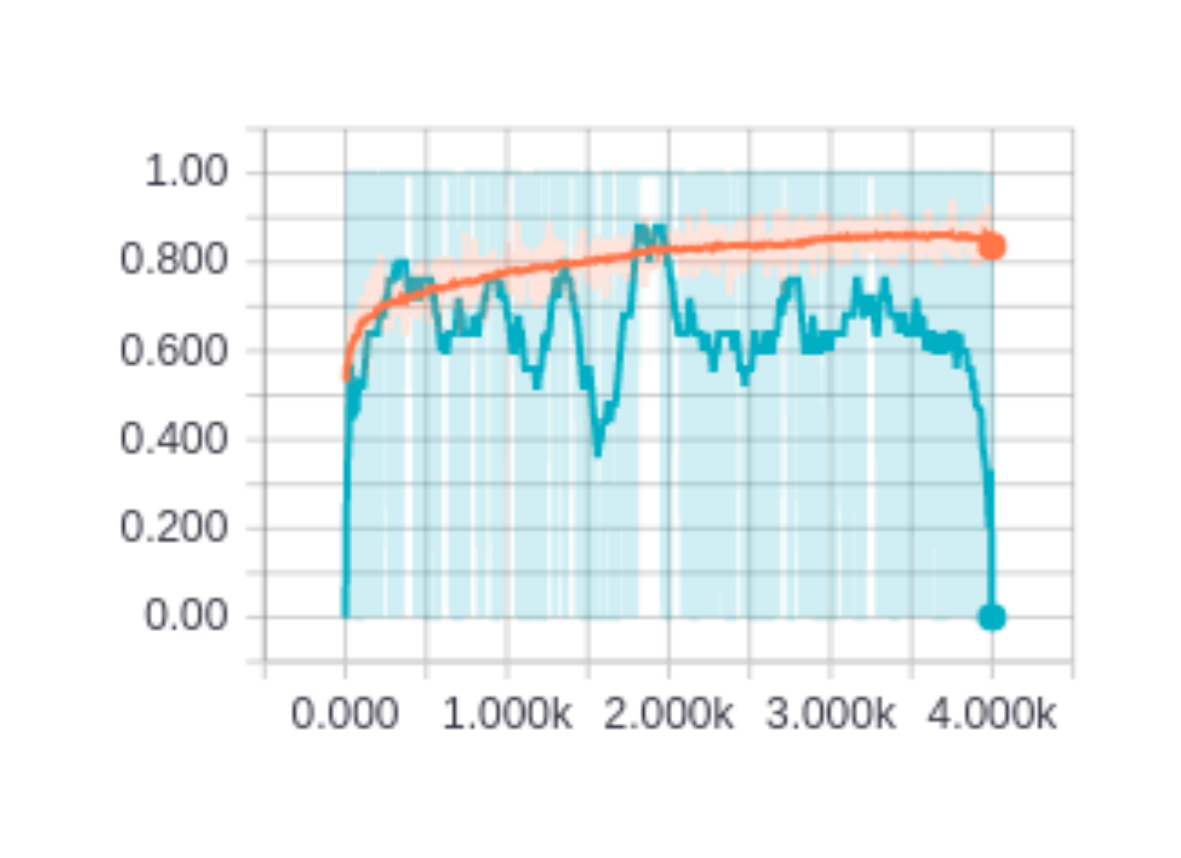}
\caption{Accuracy obtained during Training Phase} 
\label{fig2}
 \end{center}
\end{figure}

\begin{figure}[!h]
\centering
\includegraphics[width=3.2in]{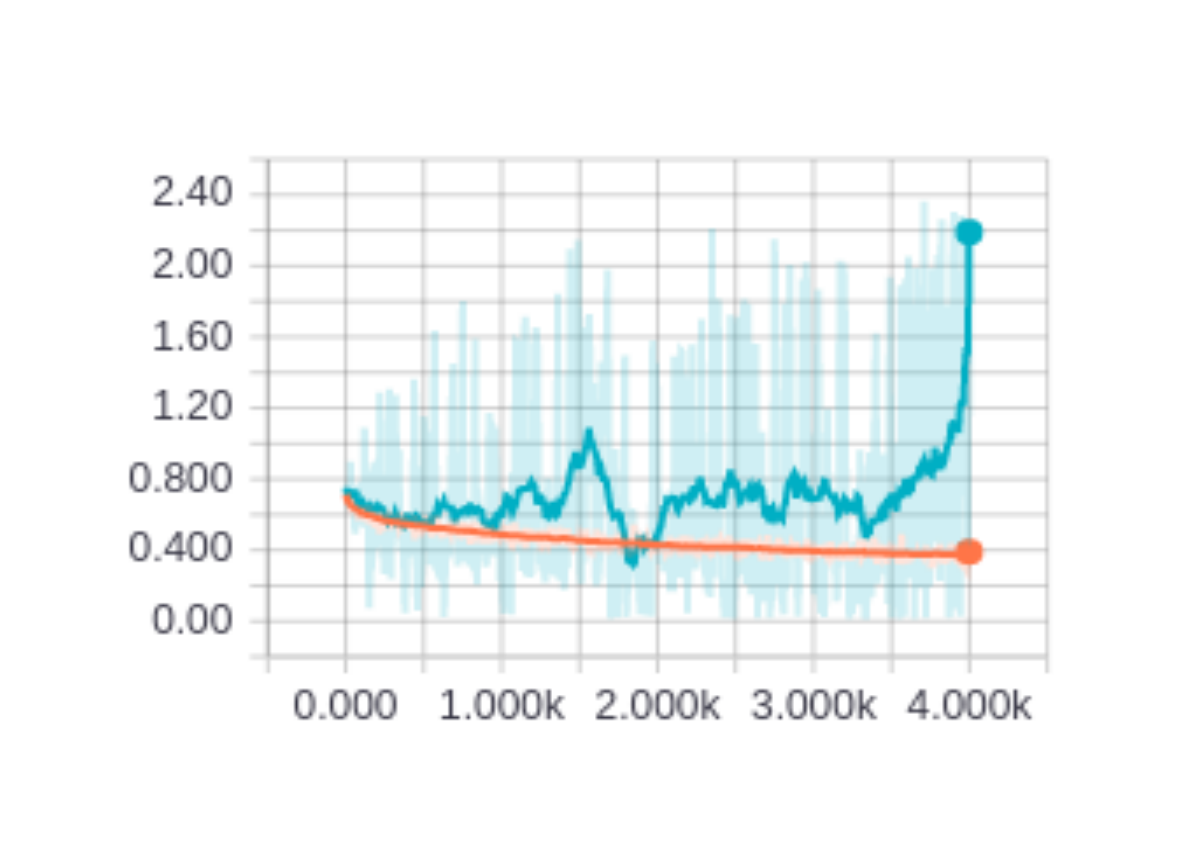}
\caption{Minimization of Loss Function}
\label{fig3}
\end{figure}

The system was trained for different number of iterations for each of the classification problems. Our system achieved high accuracy of 82.5\% and 81.4\%   at 4000th iterations in the training phase using the training images. The accuracy of the training phase has been  depicted in the figure \ref{fig2}. The minimization of the loss function during the training phase is depicted in the figure \ref{fig3}.

During the validation phase, the model scored 72\% for the phase-1 classification  and 71\% in the phase-2 and an average AUC score of 65.8\% 

\section{ Conclusion}
We have developed an automated system for skin lesion classification  which involves two different phases where the  phase-1 classifies the skin lesions either as malignant or benign  and phase-2 classifies the cause of the lesions either due to non-melanocytic  or melanocytic cells. The system classifies automatically using the image representation obtained from the dermoscopic images using Google's inspection model. We have achieved a overall AUC score of 65.8\% using the validation set provided in ISBI challenge.

\bibliographystyle{IEEEtran}
\bibliography{skin_lesion}

\begin{thebibliography}{1}
\providecommand{\url}[1]{#1}
\csname url@samestyle\endcsname
\providecommand{\newblock}{\relax}
\providecommand{\bibinfo}[2]{#2}
\providecommand{\BIBentrySTDinterwordspacing}{\spaceskip=0pt\relax}
\providecommand{\BIBentryALTinterwordstretchfactor}{4}
\providecommand{\BIBentryALTinterwordspacing}{\spaceskip=\fontdimen2\font plus
\BIBentryALTinterwordstretchfactor\fontdimen3\font minus
  \fontdimen4\font\relax}
\providecommand{\BIBforeignlanguage}[2]{{%
\expandafter\ifx\csname l@#1\endcsname\relax
\typeout{** WARNING: IEEEtran.bst: No hyphenation pattern has been}%
\typeout{** loaded for the language `#1'. Using the pattern for}%
\typeout{** the default language instead.}%
\else
\language=\csname l@#1\endcsname
\fi
#2}}
\providecommand{\BIBdecl}{\relax}
\BIBdecl

\bibitem{cnn}
\BIBentryALTinterwordspacing
C.~Szegedy, W.~Liu, Y.~Jia, P.~Sermanet, S.~Reed, D.~Anguelov, D.~Erhan,
  V.~Vanhoucke, and A.~Rabinovich, ``Going deeper with convolutions,'' in
  \emph{Computer Vision and Pattern Recognition (CVPR)}, 2015. [Online].
  Available: \url{http://arxiv.org/abs/1409.4842}
\BIBentrySTDinterwordspacing

\bibitem{Nevus}
B.~F. Viana~ACL, Gontijo~B, ``Giant congenital melanocytic nevus,'' \emph{Anais
  Brasileiros de Dermatologia}, vol.~88, no.~6, p. 863:878, 2013.

\bibitem{keratosis}
P.~RG, B.~K, R.~R, and P.~S, ``Seborrheic keratosis,'' \emph{Journal of Oral
  and Maxillofacial Pathology : JOMFP}, vol.~18, no.~2, p. 327:330, 2014.

\bibitem{color}
C.~Y. (Iris), S.~R, and U.~S. et~al., ``Skin lesion classification using
  relative color features,'' \emph{Skin research and technology : official
  journal of International Society for Bioengineering and the Skin (ISBS) [and]
  International Society for Digital Imaging of Skin (ISDIS) [and] International
  Society for Skin Imaging (ISSI)}, vol.~14, no.~1, p. 327:330, 2008.

\bibitem{CAD}
C.~W-Y, H.~A, and Y.~C.-Y. et~al., ``Computer-aided diagnosis of skin lesions
  using conventional digital photography: A reliability and feasibility
  study,'' \emph{PLoS ONE}, vol.~8, no.~11, p. e76212, 2013.

\bibitem{review}
A.~Masood and A.~A. Al-Jumaily, ``Computer aided diagnostic support system for
  skin cancer: A review of techniques and algorithms,'' \emph{International
  Journal of Biomedical Imaging}, p.~22, 2013.

\end{thebibliography}

\end{document}